\documentclass{article}

\usepackage[square, numbers]{natbib}

\usepackage[preprint]{nips_2018}

\usepackage{hyperref}       
\usepackage{url}            
\usepackage{booktabs}       
\usepackage{tabularx}
\usepackage{amsfonts}       
\usepackage{bbm}
\usepackage{nicefrac}       
\usepackage{microtype}      
\usepackage{chngpage}
\usepackage[colorinlistoftodos,prependcaption,textsize=tiny]{todonotes}
\usepackage{setspace}

\usepackage[utf8]{inputenc}
\usepackage[T1]{fontenc}
\usepackage[english]{babel}
\usepackage{dsfont}
\usepackage{amsthm,amssymb}
\usepackage{amsmath,etoolbox}
\DeclareMathOperator*{\argmax}{arg\,max}
\DeclareMathOperator*{\argmin}{arg\,min}

\usepackage{graphicx}
\usepackage[bf]{subfigure}
\usepackage{diagbox}
\usepackage[utf8]{inputenc}
\usepackage{float}
\usepackage{caption}
\usepackage[ruled]{algorithm2e}
\usepackage{algpseudocode}
\usepackage{todonotes}

\graphicspath{{./figs/}}

\title{Ranked Reward: Enabling Self-Play Reinforcement Learning for Combinatorial Optimization}

\author{
Alexandre Laterre \\
\texttt{a.laterre@instadeep.com} \\
\And Yunguan Fu \\
\texttt{y.fu@instadeep.com}
\And Mohamed Khalil Jabri \\
\texttt{mk.jabri@instadeep.com}
\And Alain--Sam Cohen \\
\texttt{as.cohen@instadeep.com} \\
\And David Kas \\
\texttt{d.kas@instadeep.com} \\
\And Karl Hajjar \\
\texttt{k.hajjar@instadeep.com} \\
\And Hui Chen \\
\texttt{h.chen@instadeep.com} \\
\And Torbj{\o}rn S. Dahl \\
\texttt{t.dahl@instadeep.com} \\
\And Amine Kerkeni \\ 
\texttt{ak@instadeep.com} \\
\And Karim Beguir \\
\texttt{kb@instadeep.com}}

\begin{document}
\maketitle

\begin{abstract}
Adversarial self-play in two-player games has delivered impressive results when used with reinforcement learning algorithms that combine deep neural networks and tree search. Algorithms like AlphaZero and Expert Iteration learn {\em tabula-rasa}, producing highly informative training data on the fly.  However, the self-play training strategy is not directly applicable to single-player games.  Recently, several practically important combinatorial optimization problems, such as the traveling salesman problem and the bin packing problem, have been reformulated as reinforcement learning problems, increasing the importance of enabling the benefits of self-play beyond two-player games. We present the Ranked Reward (R2) algorithm which accomplishes this by ranking the rewards obtained by a single agent over multiple games to create a relative performance metric. Results from applying the R2 algorithm to instances of a two-dimensional and three-dimensional bin packing problems show that it outperforms generic Monte Carlo tree search, heuristic algorithms and integer programming solvers. We also present an analysis of the ranked reward mechanism, in particular, the effects of problem instances with varying difficulty and different ranking thresholds.
\end{abstract}

\section{Introduction and Motivation}
Reinforcement learning (RL) algorithms that combine neural networks and tree search have delivered outstanding successes in two-player games such as go, chess, shogi, and hex. One of the main strengths of algorithms like AlphaZero \cite{SilverChessShogi} and Expert Iteration \cite{Anthony2017} is their capacity to learn {\em tabula rasa} through {\em self-play}. Historically, RL with self-play has also been successfully applied to the game of Backgammon \cite{Tesauro94}. Using this strategy removes the need for training data from human experts and always provides an agent with a well-matched adversary, which facilitates learning.


While self-play algorithms have proven successful for two-player games, there has been little work on applying similar principles to single-player games/problems \cite{Moerland2018}. These games include several well-known combinatorial problems that are particularly relevant to industry and represent real-world optimization challenges, such as the traveling salesman problem (TSP) and the bin packing problem (BPP).




This paper describes the {\em Ranked Reward} (R2) algorithm and results from its application to 2D and 3D BPP formulated as a single-player Markov decision process (MDP).  R2 uses a deep neural network to estimate a policy and a value function, as well as Monte Carlo tree search (MCTS) for policy improvement. In addition, it uses a reward ranking mechanism to build a single-player training curriculum that provides advantages comparable to those produced by self-play in competitive multi-agent environments.

The R2 algorithm offers a new generic method for producing approximate solutions to NP-hard optimization problems. Generic optimization approaches are typically based on algorithms such as integer programming \cite{gurobi}, that provide optimality guarantees at a high computational expense, or heuristic methods that are lighter in terms of computation but may produce unsatisfactory suboptimal solutions. The R2 algorithm outperforms heuristic approaches while scaling better than optimization solvers. We present results showing that, on 2D and 3D BPP, R2 performs better than the same deep RL algorithm without a ranked reward mechanism and also better than MCTS \citep{Browne2012}, the Lego heuristic algorithm \cite{hu2018multi} and linear programming with barrier functions \cite{gurobi}.
We also analyze the reward ranking mechanism.  In particular, we evaluate different ranking thresholds for deciding whether an episode/game should be considered a win or a loss and the effects on the overall learning process. 


In Section~\ref{sec:background} of this paper, we summarize the current state-of-the-art in deep learning for games with large search spaces. Then, in Section~\ref{sec:binpacking}, we present a single-player MDP formulation of the bin packing problem. In Section~\ref{sec:rankedreward} we describe the R2 algorithm using deep RL and tree search along with a reward ranking mechanism.  Section~\ref{sec:experiments} presents our experiments and results, and Section~\ref{sec:discussion} discusses the implications of using different reward ranking thresholds. Finally, Section~\ref{sec:conclusion} summarizes current limitations of our algorithm and future research directions.

\section{Related Work}
\label{sec:background}

Combinatorial optimization problems are widely studied in computer science and mathematics. A large number of them belongs to the NP-hard class of problems. For this reason, they have traditionally been solved using heuristic methods \cite{REGO2011427, BOYER2009658, colorni1996heuristics}. However, these approaches may need hand-crafted adaptations when applied to new problems because of their problem-specific nature.

Deep learning algorithms potentially offer an improvement on traditional optimization methods as they have provided remarkable results on classification and regression tasks \cite{Schmidhuber2015DeepLI}.
Nevertheless, their application to combinatorial optimization is not straightforward. A particular challenge is how to represent these problems in ways that allow the deployment of deep learning solutions. One way to overcome this challenge was introduced by Vinyals {et al.} \cite{Vinyals2015} through {\em Pointer Networks}, a neural architecture representing combinatorial optimization problems as sequence-to-sequence learning problems. Early Pointer Networks were trained using supervised learning methods and yielded promising results on the TSP but required datasets containing optimal solutions which can be expensive, or even impossible, to build.  Using the same network architecture, but training with actor-critic methods, removed this requirement \citep{combiRL}.

Unfortunately, the constraints inherent to the bin packing problem prohibit its representation as a sequence in the same way as the TSP.  In order to get around this, Hu {\em et al.} \cite{hu2017solving} combined a heuristic approach with RL to solve a 3D version of the problem. The main role of the heuristic is to transform the output sequence produced by the RL algorithm into a feasible solution so that its reward signal can be computed. This technique outperformed previous well-designed heuristics.

\subsection{Deep Learning with Tree Search and Self-Play}

Policy iteration algorithms that combine deep neural networks and tree search in a self-training loop, such as AlphaZero \cite{SilverChessShogi} and Expert Iteration \cite{Anthony2017}, have exceeded human performance on several two-player games. 
These algorithms use a neural network with weights $\theta$ to provide a policy $p_\theta(\cdot |s)$ and/or a state value estimate $v_\theta(s)$ for every state $s$ of the game. The tree search uses the neural network's output to focus on moves with both high probabilities according to the policy and high-value estimates. The value function also removes any need for Monte Carlo roll-outs when evaluating leaf nodes. Therefore, using a neural network to guide the search reduces both the breadth and the depth of the searches required, leading to a significant speedup. The tree search, in turn, helps to raise the performance of the neural network by providing improved MCTS-based policies during training.

Self-play allows these algorithms to learn from the games played by both players.  It also removes the need for potentially expensive training data, often produced by human experts. Such data may be biased towards human strategies, possibly away from better solutions. Another significant benefit of self-play is that an agent will always face an opponent with a similar performance level. This facilitates learning by providing the agent with just the right curriculum in order for it to keep improving \cite{openAIemergentcomplexity}. If the opponent is too weak, anything the agent does will result in a win and it will not learn to get better. If the opponent is too strong, anything the agent does will result in a loss and it will never know what changes in its strategy could produce an improvement.
The main contribution of the R2 algorithm is a relative reward mechanism for single-player games, providing the benefits of self-play in single-player MDPs and potentially making policy iteration algorithms with deep neural networks and tree search effective on a range of combinatorial optimization problems.

\section{Bin Packing as a Markov Decision Process}
\label{sec:binpacking}
The classical bin packing problem involves packing a set of items into fixed-sized bins in a way that minimizes a cost function, e.g. the number of bins required. The work presented here considers a variation of the problem where the objective is to pack a set of items into a single bin while minimizing its surface, like in the work of Hu {\em et al.} \cite{hu2017solving,hu2018multi}.


\subsection{Bin Packing Problem}
The problem involves a set of $N$ cuboid shaped items $\mathcal{I}=\{(l_i, w_i, h_i)\}_{i=1}^N$ where $l_i, w_i$ and $h_i$ denote the length, width and height of item $i$. Items can be rotated of $90^\circ$ along $x,y$ and $z$ axis, and $o_i\in\{0, 1, 2, 3, 4, 5\}$ denotes how the $i$-th item is rotated. The bottom-left-front corner of the $i$-th item placed inside the bin is denoted by $(x_i, y_i, z_i)$ with the bottom-left-front corner of the bin set to $(0,0,0)$. The problem also includes additional constraints, complexifying the environment and reducing the number of available positions in which an item can be placed. In particular, items may not overlap and an item's center of gravity needs physical support. A solution to this problem is a sequence of quintuplets $((i, x_i,y_i,z_i,o_i))_{i=1}^N$ where all items are placed inside the bin while satisfying all the constraints. The objective is then to minimize the surface of the minimal bin which contains all the items. The problem can also be reduced to 2D, where the bin is of shape $(W,H)$ and each item have only two possibilities of orientation. The objective is then to minimize the perimeter of the minimal bin.

\subsection{Markov Decision Process}
As opposed to Hu {\em et al.} \cite{hu2017solving,hu2018multi}, which address the BPPs via sequence-to-sequence methods, we formulate the problem as an MDP, where state encodes all the items and their current placement if placed and action encodes the possible positions and orientations of the unplaced items. The goal of the agent is to select actions in a way that minimizes the cost. This is reflected in the design of the reward $r_t$. As defined in Equation~\ref{eq:reward}, all non-terminal states receive a reward of $0$ while terminal states receive a reward function of the final solution's quality.
\begin{align} \label{eq:reward}
r_t = 
\begin{cases}
	\dfrac{C^*}{C}, & \text{if all items have been placed},\\
	0, & \text{otherwise},
\end{cases}
\end{align}
where costs $C$ and $C^*$ denote respectively the cost of the minimal bin at terminal state and of an ideal cube (or square) whose volume (or area) equals to the sum of the volumes (or areas) of all items. The definition of the cost $C$ of a bin $(L,W,H)$ is:
\begin{align*}
C = 
\begin{cases}
	LW+WH+LH, & \text{3D},\\
	W+H, & \text{2D}.
\end{cases}
\end{align*}

\section{The R2 Algorithm}
\label{sec:rankedreward}

When using self-play in two-player games, a funny agent faces a perfectly suited adversary at all times because no matter how weak or strong it is, the opponent always provides just the right level of opposition for the agent to learn from \cite{openAIemergentcomplexity}. The R2 algorithm reproduces the benefits of self-play for generic single-player MDPs by reshaping the rewards of a single agent according to its relative performance over recent games. A detailed description is given by Algorithm~\ref{alg:rr}.  Below we present the ranking mechanism and network model used in detail.

\subsection{Ranked Rewards}

The ranked reward mechanism compares each of the agent's solutions to its recent performance so that no matter how good it gets, it will have to surpass itself to get a positive reward.  In particular, R2 uses a size-limited buffer $\mathcal{B}$ to record recent MDP rewards and calculate a threshold value $r_\alpha$ based on a given percentile $\alpha\in (0,100)$, e.g., the threshold value $r_{75}$ is the MDP reward value closest to the $75$th percentile of the MDP rewards in the buffer.  The agent's MDP reward $R_{N-1}$ is reshaped to a \textit{ranked reward} $z\in\{\pm1\}$ according to whether or not it surpasses the threshold value: 
\begin{align}\label{eq:ranked-reward-one-th}
z = \begin{cases}
1 & r_{N-1} > r_\alpha~\text{or}~r_{N-1}=1\\
-1 & r_{N-1} < r_\alpha\\
b\sim B & r_{N-1} = r_\alpha~\text{and}~r_{N-1} < 1
\end{cases},
\end{align}
where $z=b$ is sampled from a binary random variable $B$ such that when $r=r_\alpha$, $z$ equals to $\pm1$ randomly. This way, the player is provided with samples of recent games labeled relatively to the agent's current performance, providing information on which policies will improve its present capabilities. The random variable is used to break ties and assure constant-learning. Indeed, if we set $z$ to $1$ when $r=r_\alpha<1$, the agent does not have an incentive to beat the threshold since it can obtain positive rewards by staying at its current performance level. The ranked rewards are then used as targets for the value estimation neural network and as the value to backup for terminal nodes during MCTS.

\begin{algorithm}[t]
\SetAlgorithmName{Algorithm}{}
\KwData{}
\textbf{Input}: a percentile $\alpha$ and a mini-batch size $b$

Initialize fixed size buffers $\mathcal{D} = \{ \}$ and $\mathcal{B} = \{ \}$

Initialize parameters $\theta_0$ of the neural network, $f_{\theta_0}$

\For{$k = 0, 1, \ldots$}{
    
    \For{\emph{episode = 1}, \dots, M}{
    	Sample an initial state $s_0$
        
        \For{\emph{t = 0}, \ldots, N-\emph{1}}{
        	Perform a Monte Carlo tree search consisting of $S$ simulations guided by $f_{\theta_k}$
            
            Extract MCTS-improved policy $\pi(\cdot\vert s_t)$
            
            Sample action $a_t \sim \pi(\cdot\vert s_t)$
            
            Take action $a_t$ and observe new state $s_{t+1}$
        }
        
        Compute MDP reward $r_{N-1}$ and store it in $\mathcal{B}$
        
        Compute threshold $r_\alpha$ based on the MDP rewards in $\mathcal{B}$
        
        Reshape to ranked reward $z$ as explained in \eqref{eq:ranked-reward-one-th}
        
        Store all triplets $(s_t, \pi(\cdot \mid s_t), z)$ in $\mathcal{D}$ for $t=0, \ldots, N-1$
    }  
    
    $\theta \leftarrow \theta_k$
    
    \For{\emph{step = 1}, \ldots, J}{
    	Sample mini-batch $\mathcal{J}$ of size $b$ uniformly from from $\mathcal{D}$
        
        Update $\theta$ by performing one optimization step using mini-batch $\mathcal{J}$
    }
    
    $\theta_{k+1} \leftarrow \theta$
    
}
\caption{R2}
\label{alg:rr}
\end{algorithm}

\subsection{Neural Network Architecture}
The input of the neural network consists of the set of feasible actions where each action consists of features describing the chosen item, id and orientation, as well as the placement location.  The network architecture satisfies two critical requirements. First, it is \textit{permutation invariant}, i.e. any permutation of the input set results in the same output permutation.  Second, the model is able to process input sets of any size since the size of the available action space varies as the items are being placed.

Each action in the action space is fed independently into a feed-forward network taking fixed-size inputs.  The resulting feature-space embeddings are aggregated using pooling operations. The final output is obtained by combining these with the embeddings through further non-linear processing to obtain the agent's policy and its state-value estimate.

\section{Experiments and Results}
\label{sec:experiments}
To validate our approach and evaluate its effectiveness, we first considered 2D and 3D BPP with $10$ items.  The items were generated by repeatedly dividing a square/cube of size $10$.  The process for generating items randomly is presented in detail in Appendix~\ref{subsec:generation-of-bpp}.  The training process is as follows: at each iteration, $50$ problems are generated and solved by R2 with a reward buffer of size $250$, used to define the threshold $r_\alpha$.  All MCTS instances perform $300$ simulations per move.  The reshaped rewards alongside the MCTS-improved policies are stored in a dataset and used during training.  The neural network is trained by gradient descent (Adam) using mini-batches of size $32$, uniformly sampled from the last $500$ games.  At each iteration, $50$ steps of gradient descent are performed.
Each experiment ran on a NVIDIA Tesla V100 GPU card for up to two days.

\subsection{Ranked Reward with Different Thresholds}
\begin{figure}[ht!]
\begin{center}
\subfigure[Mean rewards for 2D BPPs]{\includegraphics[width=0.45\textwidth]{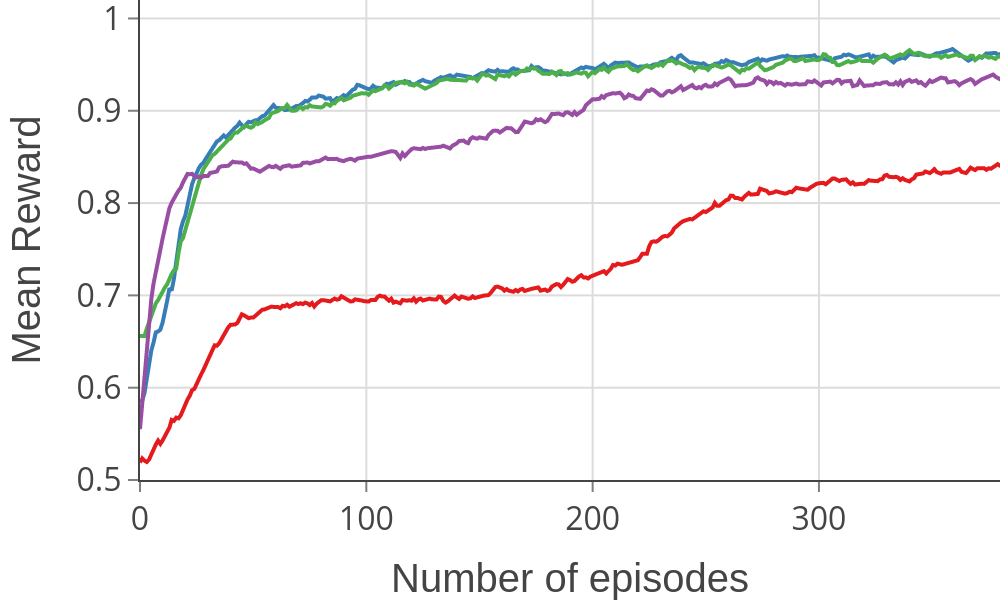}\label{fig:bpp2d_reward}}
    \subfigure[Optimality percentages for 2D BPPs]{\includegraphics[width=0.45\textwidth]{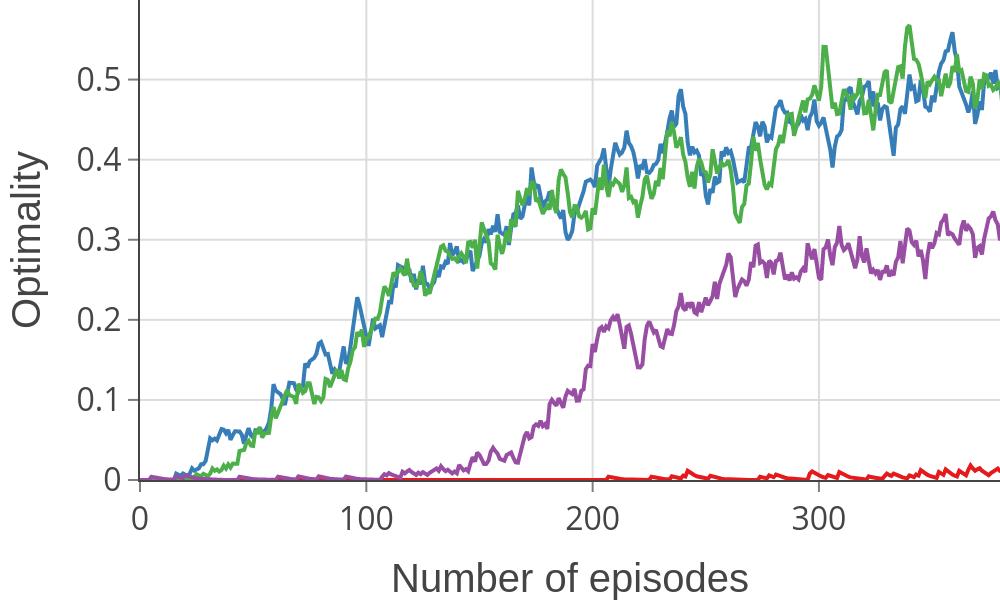}\label{fig:bpp2d_optimality}}
    \subfigure[Mean rewards for 3D BPPs]{\includegraphics[width=0.45\textwidth]{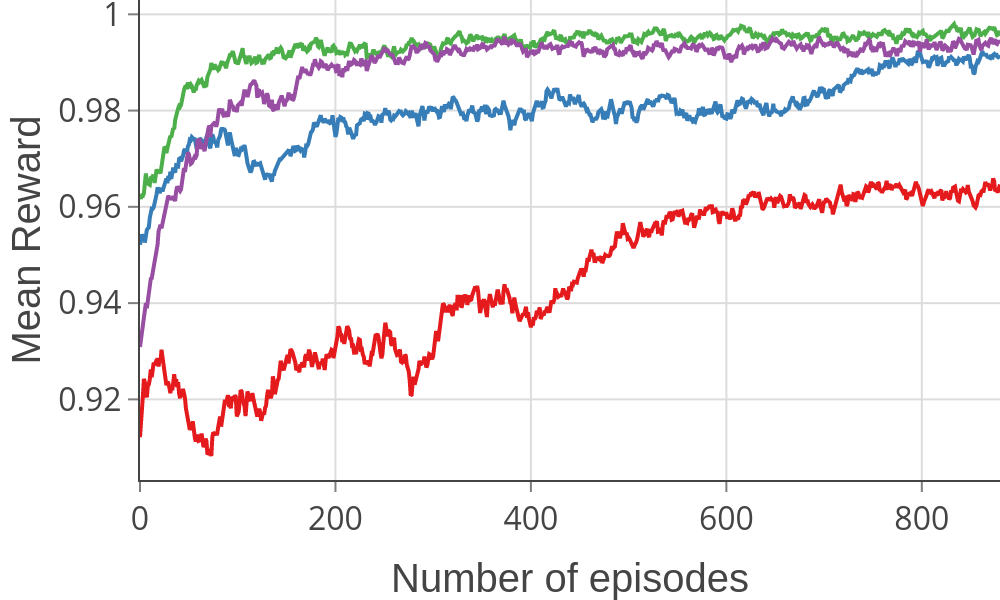}\label{fig:bpp3d_reward}}
    \subfigure[Optimality percentage for 3D BPPS]{\includegraphics[width=0.45\textwidth]{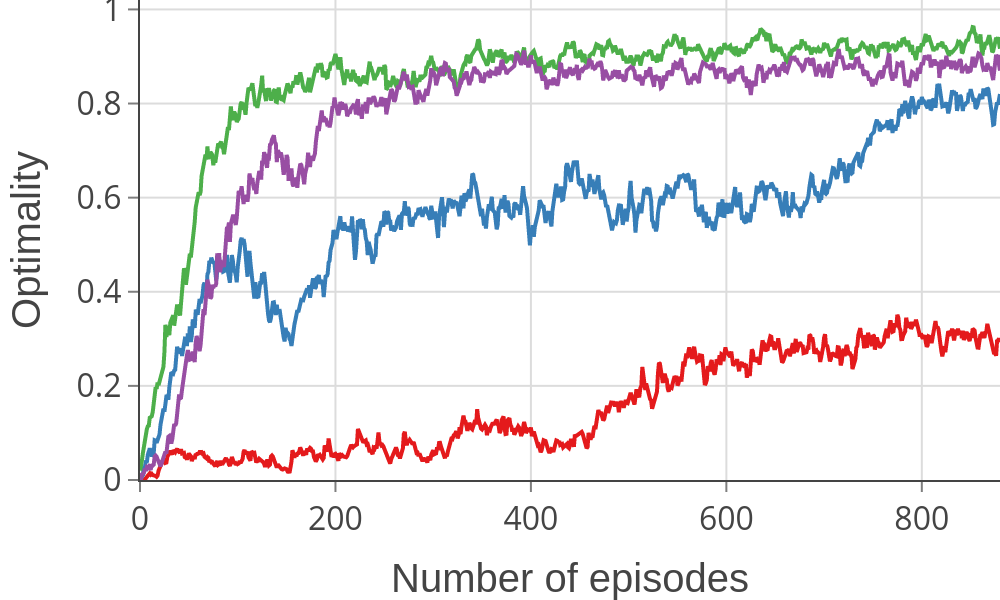}
    \label{fig:bpp3d_optimality}}
    \caption{\small Mean rewards and optimality percentages of R2 on 2D and 3D bin packing problems with percentile of 50 (blue), 75 (green), 90 (purple) and Rank-Free (red).}
    \label{fig:bpp-threshold}
\end{center}
\end{figure}

We first compared the performance of the R2 algorithm with three different $\alpha$-percentiles: $50$, $75$ and $90$. We also included a version using the MDP reward directly without ranking as the target for the value function estimate.
The learning curves are presented in Figure~\ref{fig:bpp-threshold} for games with 10 items in 2D and 3D.  R2 with an $\alpha$-percentile of $75\%$ achieved the best training performances in both cases.  Regardless of the threshold, R2 outperforms Rank-Free in both mean rewards and optimality percentages with a faster convergence. 
We later evaluated the different threshold on problems with 20, 30 and 50 items.  On these larger problems, the $50\%$ threshold produced the best final performance.  The details of these evaluations are given in Appendix~\ref{app:thresholds}.

\subsection{Comparison with Other Methods}
\label{2}

To evaluate the performance of R2, we compared Rank-$75\%$ with a set of baselines as comparison: trained neural network agent without MCTS; a supervised agent; a plain MCTS agent using Monte-Carlo rollouts for state-value estimation~\cite{Browne2012}; the Lego heuristic search algorithm~\cite{hu2017solving,hu2018multi}; and a solver agent using integer programming~\cite{gurobi} (see appendix \ref{app:benchmark} for further details). We used $100$ games as the test set for all the algorithms.



\begin{figure}
\centering
\subfigure[2D BPPs with a total area of items equals to $900$]{\includegraphics[width=0.89\textwidth]{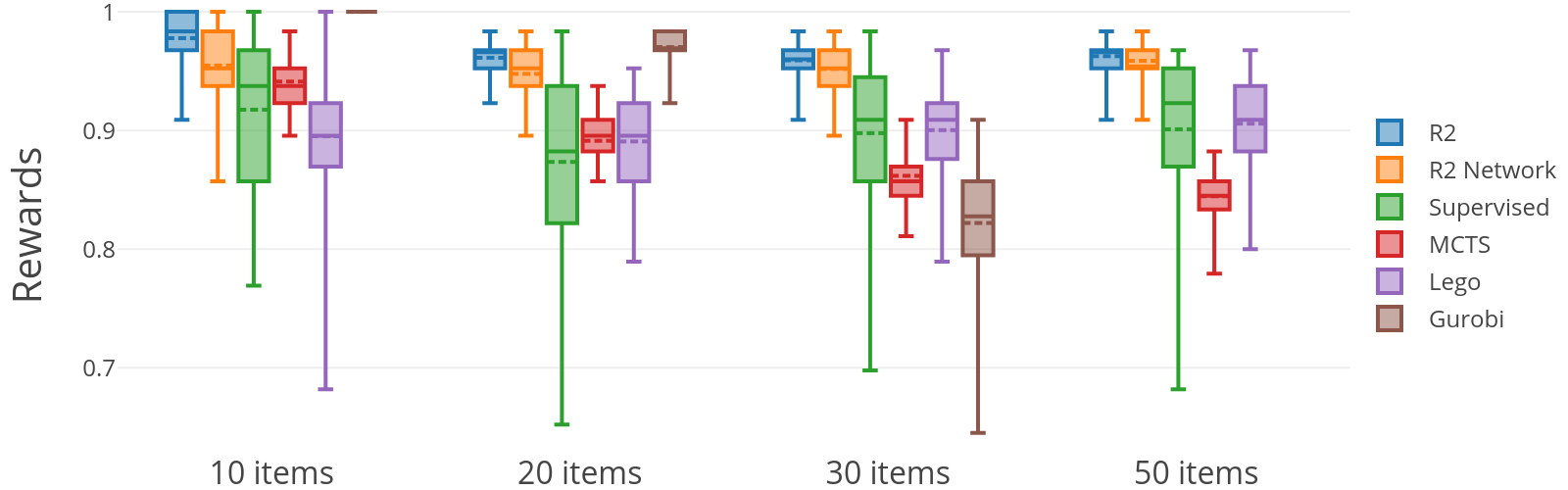}\label{fig:2d_box}}
\subfigure[3D BPPs with a total volume of items equals to $27000$]{\includegraphics[width=0.89\textwidth]{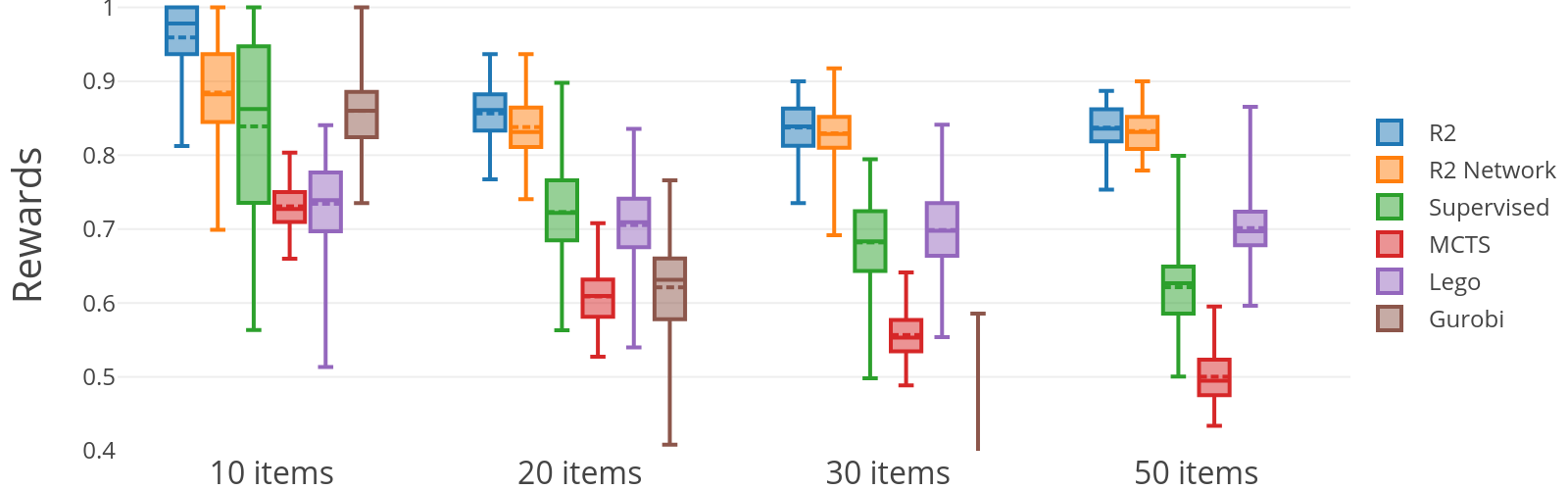}\label{fig:3d_box}}
\caption{Performance on 2D and 3D games for Rank-$75\%$ and other algorithms.}
\label{fig:box_plot}
\end{figure}

As shown in Figure~\ref{fig:2d_box} and Figure~\ref{fig:3d_box}, R2 always outperformed its alternatives. Especially, we can observe that the trained network performed at least as good as the pure MCTS algorithm and Lego heuristic, which proves that the architecture of the neural network is well designed and it is capable of learning good policies. 
With the help of specifically designed training data, the supervised agent achieved also a superior performance compared to the pure MCTS and Lego heuristic on small instances but the performance decreased quickly as the size of problems increased. Concerning the Gurobi solver, it was given five minutes, around the same amount of time that R2 algorithm used. The solver was able to find optimal solutions for problems with $10$ items, but for problems with more items, sometimes feasible solutions could barely be found. Hence the rewards vanished to zero (which are not shown in the figures). Note that the algorithm performed better in 2D with $50$ items since the total volume is fixed and the items sizes are reduced. Therefore the problems become simpler.

\begin{figure}[ht]
\centering
\subfigure[Lego]{\includegraphics[trim={0 0 400px 450px},clip,width=0.18\textwidth]{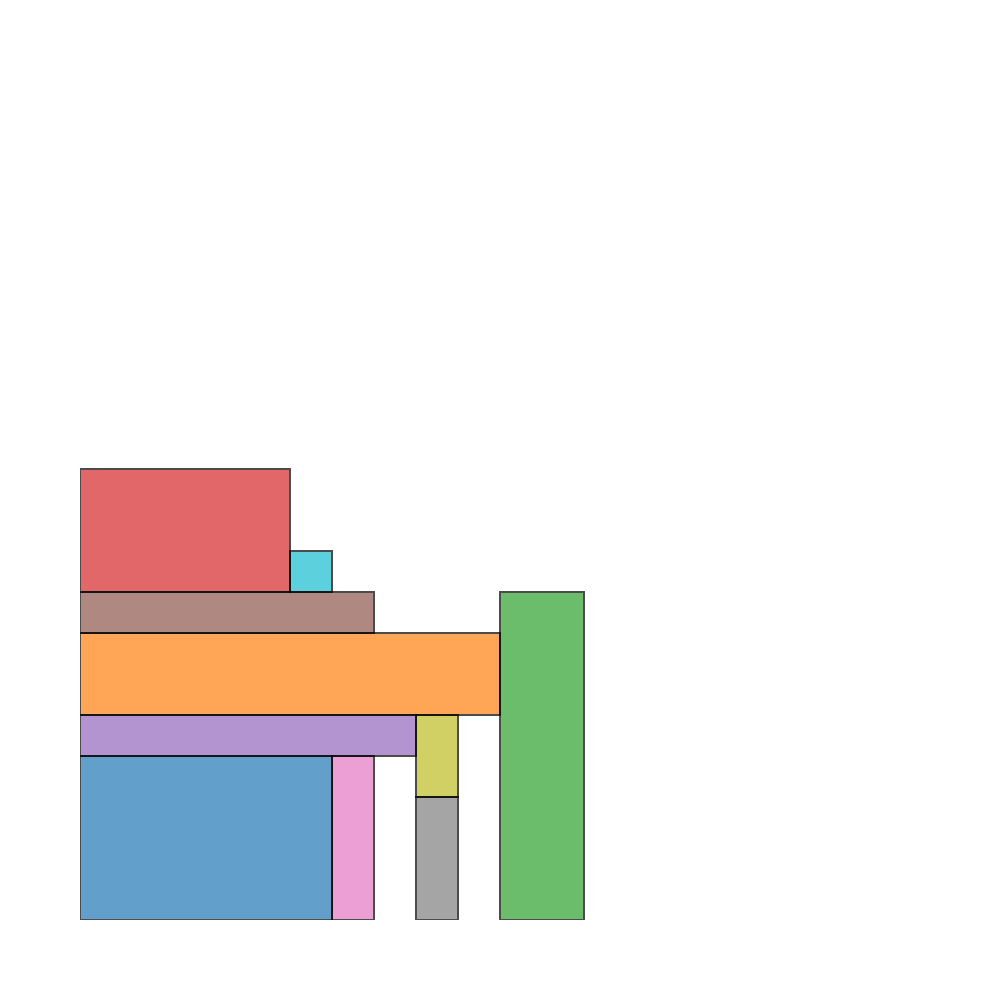}\label{fig:bpp-visu-2d-lego}}
\subfigure[Rank-$75\%$]{\includegraphics[trim={0 0 400px 450px},clip,width=0.18\textwidth]{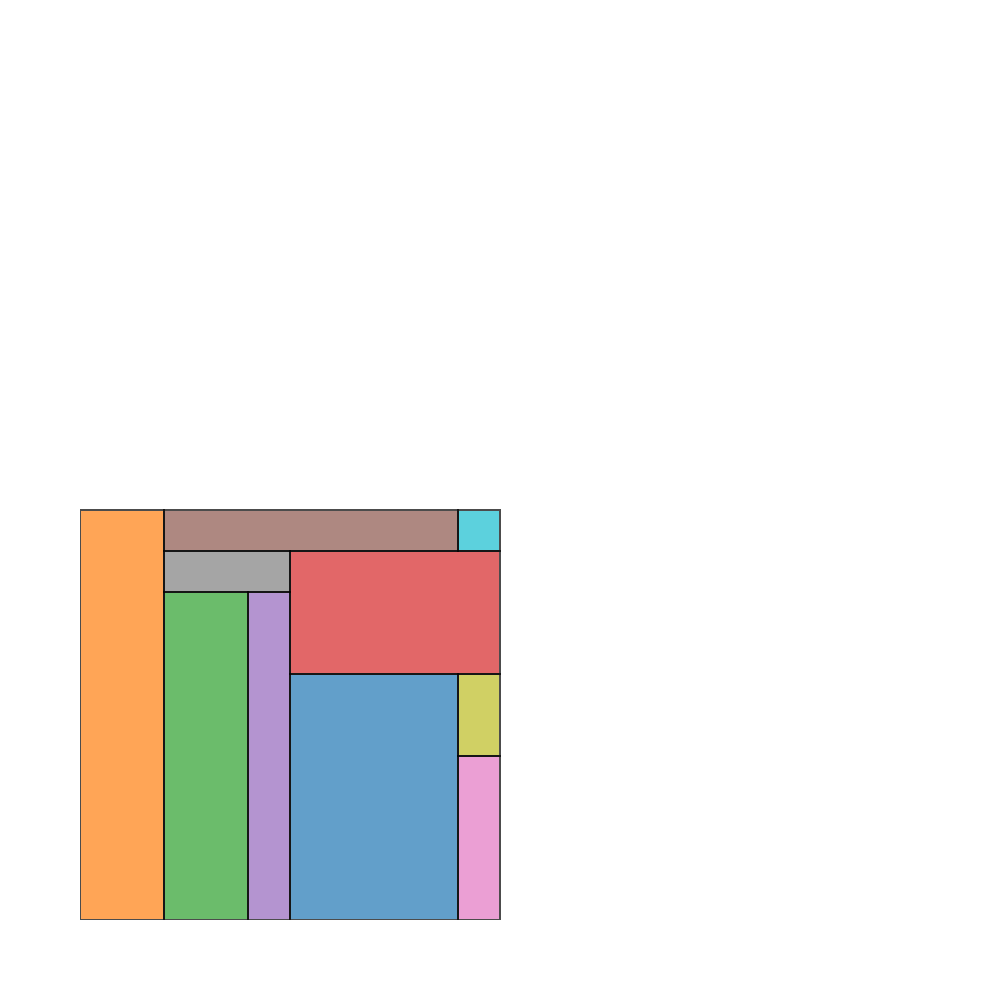}\label{fig:bpp-visu-2d-r2}}
\subfigure[Lego]{\includegraphics[width=0.18\textwidth]{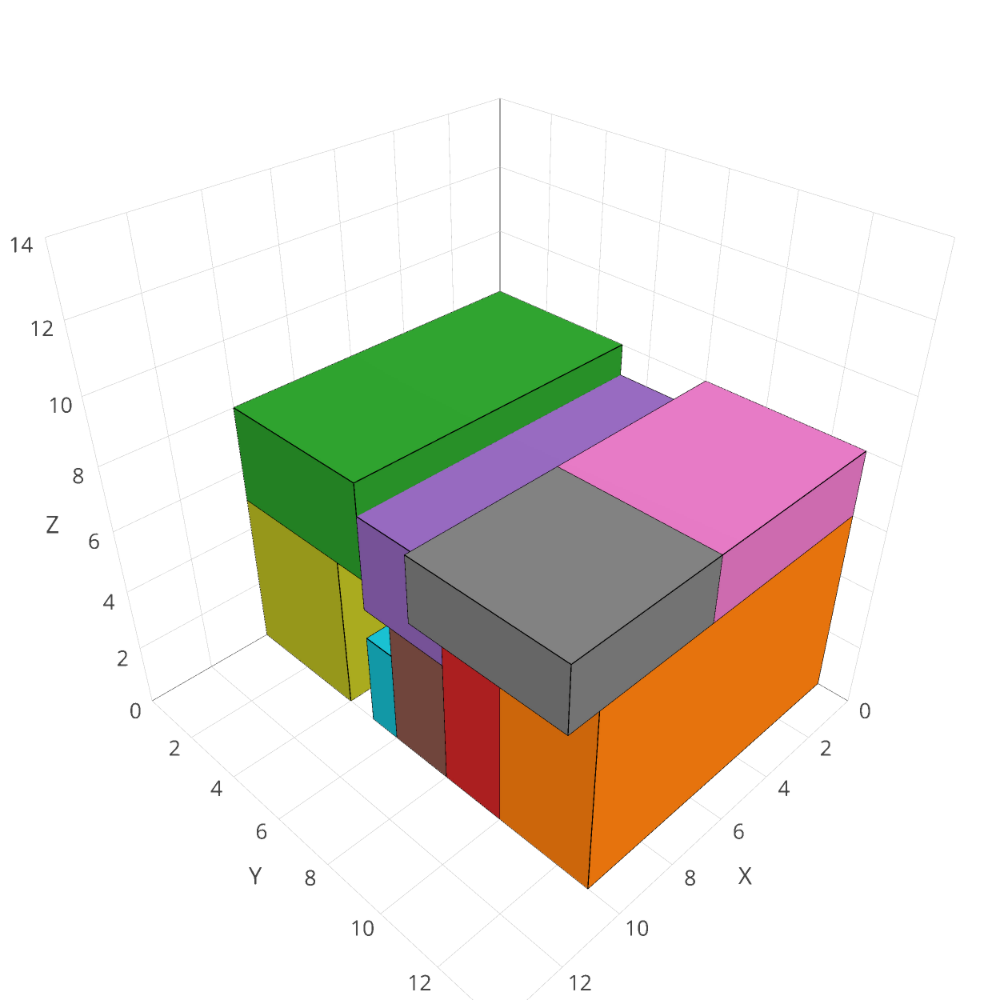}\label{fig:bpp-visu-3d-lego}}
\subfigure[Rank-$75\%$]{\includegraphics[width=0.18\textwidth]{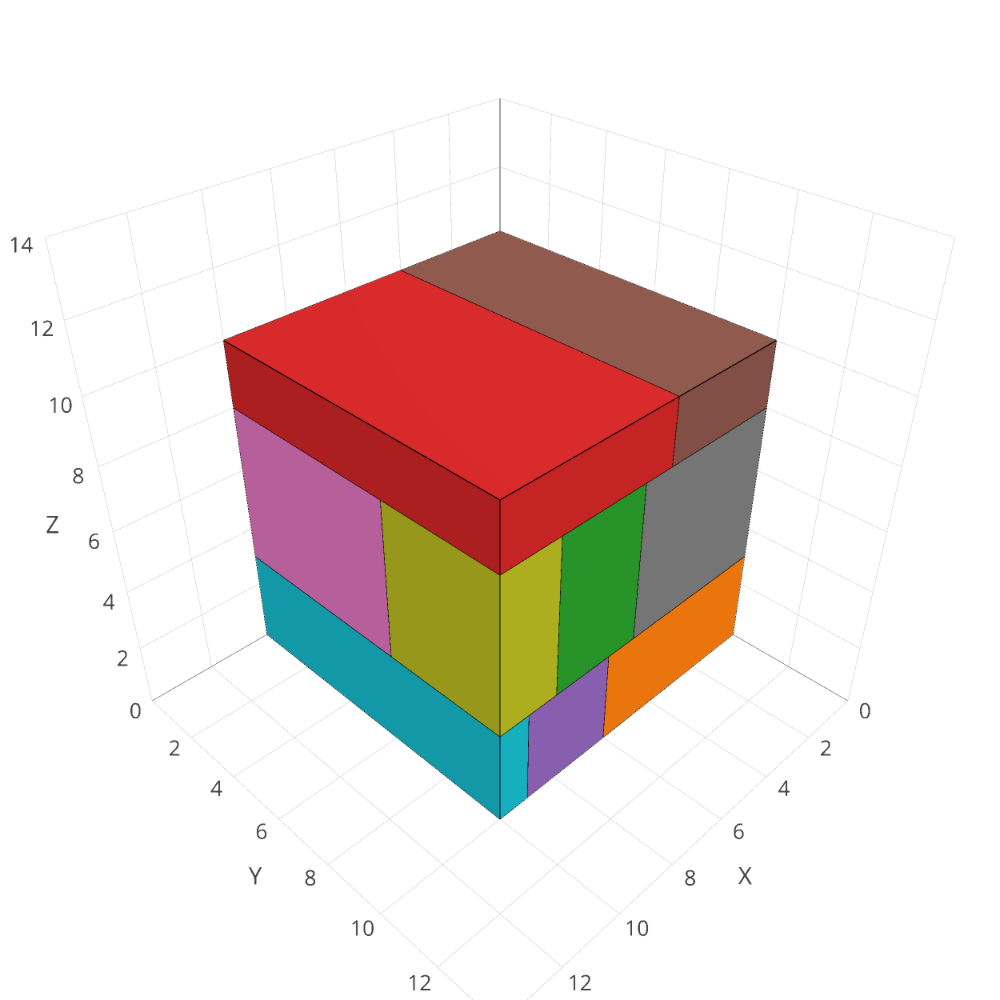}\label{fig:bpp-visu-3d-r2}}
\caption{\small Visualization of the solution by Lego and Rank-75\% in 2D and 3D.}
\label{fig:bpp-exp-test-2dn10}
\end{figure}

Furthermore, we tested the trained network with different percentiles on the same set of problems to compare the generalization ability. The results are presented in Appendix~\ref{app:thresholds} and we found that although Rank-$75\%$ achieved better training performance, Rank-$50\%$ was more robust and had even a better test performance. For illustration, we provide in Figure~\ref{fig:bpp-exp-test-2dn10} a visualization of solutions given by Lego and Rank-75\% in 2D and 3D.

 \vspace*{-0.2cm}

\section{Discussion}
\label{sec:discussion}

In this section, we present our analyzes of two critical facets of learning with ranked rewards.  The first if the issue of constructing an effective ranking when the agent plays game instances with different level of difficulty.  The second is the issue of identifying an optimal threshold and analyzing the potential benefits of, and problems with, high and low ranking thresholds.  

\subsection{Ranking the Performance on Games of Different Difficulty}

When using self-play in two-player games, the sequence of actions which lead to victory is superior to the sequence leading to a loss.  This pushes the agent's policy in the direction of the winner's actions.  However, when using ranked reward, there is no direct relationship between the sequences of actions of the agent on two different games.  The agent might have performed poorly on one game because the difficulty level of that particular game was higher than most other games.  This effect introduces significant amounts of noise to the reward signal. 

A workaround would be to include the difficulty level in the ranking process, such that poor game outcomes still obtain good rankings for complex game instances. However, this approach would require access to the difficulty level of the games, which is unlikely to happen. A second approach would incorporate randomness, e.g. additional noises into the prior distribution of the policy neural network, into the agent's behavior while making it play the same game multiple times. By this mechanism, we can assure comparability of the game outcomes and correlations between these and the agents' policies. We tried the second approach on the bin packing problem but no further improvement was noticed due to the relative similarity of the difficulty level of the games\footnote{The empirical reward distribution of the Lego heuristic corresponds to a bell-shape with a small standard deviation, which suggests that different games have relatively the same difficulty level.}.
The Ranked Reward algorithm as described in this paper is therefore appropriate for problems in which different problem instances have relatively the same difficulty level. For complex problems in which difficulty could vary widely from one instance to another, R2 would certainly benefit from applying the mechanism above to ensure comparability in the game outcomes.

\subsection{The Effects of Ranking Thresholds on Learning}

\begin{figure}[ht]
\centering
\subfigure[Rank-Free]{\includegraphics[width=0.45\textwidth]{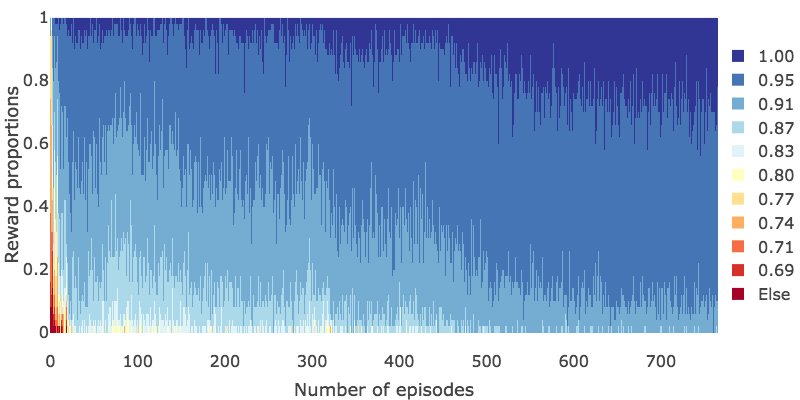}\label{fig:barno}}
\hfill
\subfigure[Ranked-$50\%$]{\includegraphics[width=0.45\textwidth]{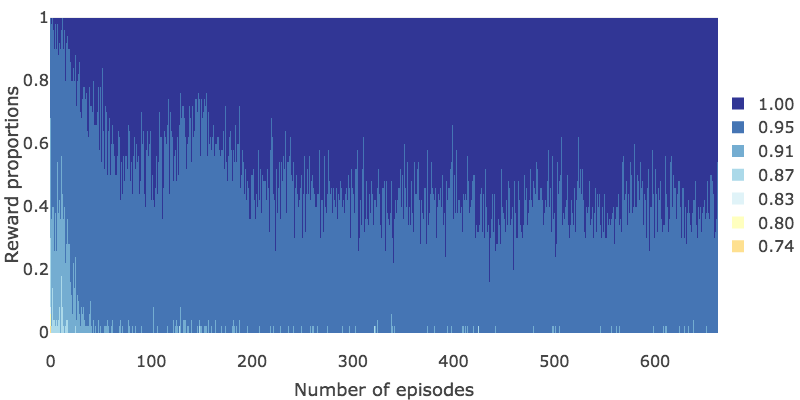}\label{fig:bar50}}
\subfigure[Ranked-$75\%$]{\includegraphics[width=0.45\textwidth]{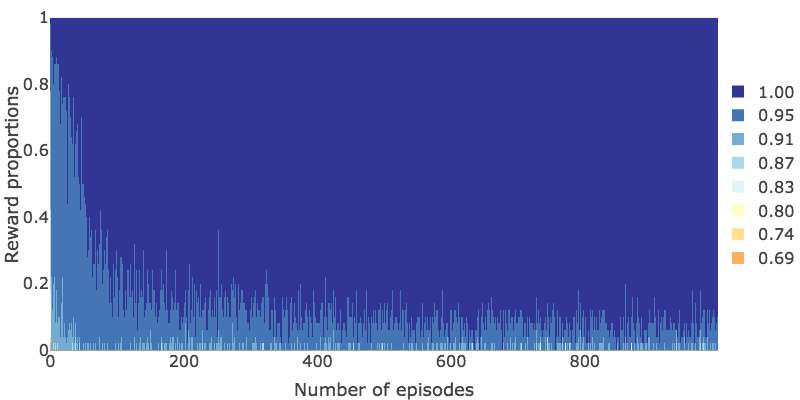}\label{fig:bar75}}
\hfill
\subfigure[Ranked-$90\%$]{\includegraphics[width=0.45\textwidth]{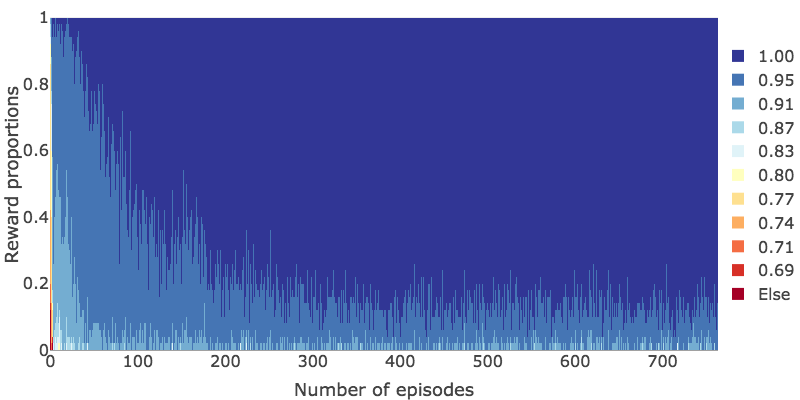}\label{fig:bar90}}
\caption{\small Evolution of the reward proportions with different percentiles $\alpha$ during training in 2D BPPs. Dark blue denotes the maximum reward of $1$ and red denotes the minimum reward of $0$. Ranked (75\%) achieves better performance and stability than others.}
\label{fig:reward_proportion}
\end{figure}

Although the performances of R2 are relatively robust to the percentile used, we studied the difference of learning performances when using other percentile values. 
We analyzed and interpreted the evolution of the reward distribution over time for four different cases, rank-free and $\alpha$ values of $50\%$, $75\%$ and $90\%$ on 2D BPP with 10 items, as shown in Figure~\ref{fig:reward_proportion}. 


Figures~\ref{fig:barno} and~\ref{fig:bar50} show that, although better solutions dominate, optimal solutions are not picked up convincingly in the {rank-free} and {ranked $50\%$} cases. In the {rank-free} case, the difference between the optimal MDP reward $1.0$ and the nearest sub-optimal MDP reward $0.96$ is relatively small, potentially making it difficult to distinguish between them effectively. In the {ranked $50\%$} case, Figure~\ref{fig:bar75}, all the games in the top half of the buffer receive a ranked reward of $1.0$.  This provides positive feedback to a significant number of sub-optimal games and it is potentially this effect that makes the convergence slow in this case. In the {ranked $75\%$} case, only the top quarter of the buffer receive a ranked reward of $1.0$. This effectively picks up the optimal solutions and expels sub-optimal solutions from the buffer almost completely. The {ranked $90\%$} case, Figure~\ref{fig:bar90}, also converges quickly, but less quickly than the {ranked $75\%$} case.  In this case, a smaller number of good solutions will receive positive feedback, giving the learning a much smaller reward signal initially. This could be the cause of the slower convergence rate, though optimal solutions always receive a positive reward. 

The impact of the percentile $\alpha$ on the performance follows our intuitive understanding of human learning. Setting the threshold at $50\%$ is equivalent to making the agent play against an opponent of the same level, as it has a predetermined $50\%$ chance of winning. Increasing the percentile value corresponds to improving the opponent's level, as it makes it harder to obtain a reward of $1$. In our context, when the percentile changes from $50\%$ to $75\%$, the probability of winning falls to $25\%$.
In general, higher thresholds lead to faster learning, i.e., the proportion of high-reward games increases faster. However, Figure~\ref{fig:bar90} shows that, for a threshold of $90\%$, a larger residue of low-reward games remains. Although this effect is stronger for 3D problems. These instabilities could explain the weaker final performance of the higher thresholds. To explain this, we can hypothesize that if the opponent is too strong, the learning process will suffer as the agent can very rarely affect the game outcome even when it plays significantly better than its current mean performance level.

\section{Conclusion and Future Work}
\label{sec:conclusion}

The results presented in this work show that R2 outperforms the selected alternatives both on the 2D and 3D bin packing problems with 10, 20, 30 and 50 items. In particular, the capacity of the algorithm to outperform its competitors in large instances makes it suitable for solving real-life problem instances.

By ranking the rewards obtained over recent games, R2 provides a threshold-based relative performance metric.  This enables it to reproduce the benefits of self-play for single-player games, removing the requirement for training data and providing a well-suited adversary throughout the learning process. Consequently, R2 outperforms the selected alternatives as well as its rank-free counterpart, improving on the performance of the best alternative, the Gurobi solver, by more than $6\%$ when using a threshold value of $75\%$. As the number of items grows, the difference can reach up to $15\%$. An analysis of the effects of different percentiles $\alpha$ has shown that higher thresholds perform better up to a point after which learning becomes unstable and performance decreases.

For now, our implementation of the bin packing problem only considers problems that do not contain any spare space, i.e., square packings with no gaps. Even though this helps us to evaluate the algorithm's performance, it introduces an undesirable bias. Future research should evaluate the algorithm on a wider range of problems, for which the optimal solution is unknown and not necessarily square.

The R2 algorithm is potentially applicable to a wide range of optimization tasks, though it has so far been used only on the bin packing. In the future, we will consider other optimization problems such as the Traveling Salesman Problem to further evaluate its effectiveness.




\newpage
\small
\bibliography{insta}
\bibliographystyle{plain}

\newpage
\appendix
\section{Generation of Bin Packing Problems}
\label{subsec:generation-of-bpp}

Due to the lack of existing datasets for the bin packing problems, we generated instances artificially via splitting a cube (or square) randomly. The detailed algorithm is given in Algorithm \ref{algo:bpp-generation}. Especially, the information of optimal solution is not used in the definition of the problem and the definition of the final reward is applicable for problems without the knowledge of the optimal solution.

\begin{algorithm}
\DontPrintSemicolon 
\SetFuncSty{textproc} 
\SetKwInOut{Input}{Input}
\SetKwInOut{Output}{Output}
\Input{Number of item $N$, size of optimal bin $(L,W,H)$, random seed $s$.}
\Output{Item set $\mathcal{I}$}
\BlankLine
\SetKwFunction{FMain}{$\textproc{BPP Generator}$}
\SetKwProg{Pn}{Function}{:}{\Return{$\mathcal{I}$}}
\Pn{\FMain{$N,L,W,H,s$}}{
	Initialize the items list $\mathcal{I} = \{(L,W,H)\}$.\;
    \While{\upshape $|\mathcal{I}|<N$} {
    	Pop an item randomly from $\mathcal{I}$ by the item's volume.\;
        Choose an axis randomly by the length of edge.\;
        Choose a position randomly on the axis by the distance to the center of edge.\;
        Split the item into two and add them into $\mathcal{I}$.\;
    }
}
\caption{Bin Packing Problem Generator}
\label{algo:bpp-generation}
\end{algorithm}


\section{Benchmark Algorithms}
\label{app:benchmark}

\begin{algorithm}
\DontPrintSemicolon 
\SetFuncSty{textproc} 
\SetKwInOut{Input}{Input}
\Input{Items $\mathcal{I}=\{(l_i,w_i,h_i)\}_{i=1}^N$.}
\BlankLine
\SetKwFunction{FMain}{$\textproc{Lego}$}
\SetKwProg{Pn}{Function}{:}{\Return{}}
\Pn{\FMain{$\mathcal{I}$}}{
    \For{\upshape $t \gets 0$ \textbf{to} $N-1$} {
  		\uIf{$t=0$}{
    		Choose the item of largest volume, i.e. $i_0=\argmax\limits_il_iw_ih_i$.\;
            Rotate it such that $l_{i_0} \geq w_{i_0} \geq h_{i_0}$ and place it at $(0,0,0)$.\;
  		}
  	    \Else{
        	Select the item of the action which minimizes the percentage of the wasted space in the minimal bin, i.e. $i_t=\argmin\limits_{(i, x, y, z, r)}\frac{V_{\text{placed items}}}{V_{\text{bin}}}$.\;
      		Select the action which minimized the surface, i.e. $(x_{i_t}, y_{i_t}, z_{i_t}, r_{i_t})=\argmin\limits_{(i_t,x,y,z,r)}S_\text{bin}$.\;
            Perform the action $(i_t,x_{i_t}, y_{i_t}, z_{i_t}, r_{i_t})$.\;
  		}
}
}
\caption{Lego Heuristic Algorithm (3D)}
\label{algo:bpp-lego-3d}
\end{algorithm}

\textbf{Details of the benchmark algorithms}
\begin{itemize}
\item {\bf Plain MCTS} The plain MCTS agent used $300$ simulations per move just like R2 and executed a single Monte Carlo roll-out per simulation to estimate state values.
\item {\bf Lego Heuristic}  The Lego algorithm worked sequentially by first selecting the item minimizing the wasted space in the bin, and then selecting the orientation and position of the chosen item to minimize the bin size.
\item {\bf Supervised Learning} The BPP instances were generated with a known optimal solution for each problem, we designed a Lego-like heuristic algorithm defining a corresponding optimal sequence of actions $\{a_i\}_{0\leq i\leq N-1}$ for such solution. We used the state-action pairs $(s_i,a_{i})$ to train the policy-head of the R2 neural network as a one-class classification problem: given state $s_i$.
\item {\bf Gurobi Solver} With exactly the same constraints as reinforcement learning environments, we wrote two mathematical models for 2D and 3D problems and used Gurobi solver to solve them. Precisely, all constraints are linear and the objective is linear for 2D and quadratic for 3D. Solver were given $8$ CPUs and five minutes in total (roughly the same amount of time as R2 agents with MCTS) and it reported the best found feasible solution. 
\end{itemize}

\section{Comparison of the Results for Different Threshold Values}
\label{app:thresholds}
Here we present the performance of the networks trained with the R2 algorithm, without doing MCTS simulations.

\begin{table*}[ht!]
\begin{center}
\begin{sc}
\begin{tabular}{l|c|c|c|c|}
\toprule
          \diagbox[width=8em,trim=l]{Algo}{Items} & 10 & 20 & 30 & 50 \\ \midrule
          
Rank-Free  &  $0.938 (\pm 0.030)$  & $0.946(\pm0.020)$ & $0.952(\pm 0.021)$ & $0.960 (\pm0.013)$  \\   
Rank-$50\%$ & $\boldsymbol{0.953 (\pm 0.027)}$ & $\boldsymbol{0.948(\pm 0.020)}$ & $\boldsymbol{0.954(\pm 0.015)}$ & $\boldsymbol{0.960(0.016)}$ \\
Rank-$75\%$ & $0.926 ( \pm 0.064)$ & $0.933(\pm0.044)$ & $0.940(\pm0.040)$ & $0.952(\pm 0.028)$  \\
Rank-$90\%$ & $0.950(\pm 0.037)$ &$0.944(\pm 0.024) $ &$0.948(\pm 0.027)$ & $0.958(\pm 0.016)$ \\
\bottomrule
\end{tabular}
\end{sc}
\end{center}
\vskip -0.1in
\caption{Mean rewards on 2D BPPs with a total area of items equals to $900$.}
\label{tab:bpp-exp-test}
\end{table*}

\begin{table*}[ht!]
\begin{center}
\begin{sc}
\begin{tabular}{l|c|c|c|c|}
\toprule
          \diagbox[width=8em,trim=l]{Algo}{Items} & 10 & 20 & 30 & 50 \\ \midrule
Rank-Free  &  $0.807 (\pm 0.061)$  & $0.769(\pm0.052)$ & $0.770(\pm 0.039)$ & $0.762 (\pm0.037)$  \\   
Rank-$50\%$ & $0.902 (\pm 0.058)$ & $\boldsymbol{0.850(\pm 0.032)}$ & $\boldsymbol{0.844(\pm 0.030)}$ & $\boldsymbol{0.838(0.034)}$ \\
Rank-$75\%$ & $\boldsymbol{0.903 ( \pm 0.060)}$ & $0.840(\pm0.030)$ & $0.815(\pm0.044)$ & $0.816(\pm 0.041)$  \\
Rank-$90\%$ & $0.871(\pm 0.060)$ &$0.832(\pm 0.034) $ &$0.815(\pm 0.044)$ & $0.817(\pm 0.037)$ \\
\bottomrule
\end{tabular}
\end{sc}
\end{center}
\vskip -0.1in
\caption{Mean rewards on 3D BPPs with a total volume of items equals to $27000$.}
\label{tab:ranked-reward-vs-others}
\end{table*}

\begin{figure}[ht!]
\centering
\subfigure[2D BPPs with a total area of items equals to $900$]{\includegraphics[width=0.89\textwidth]{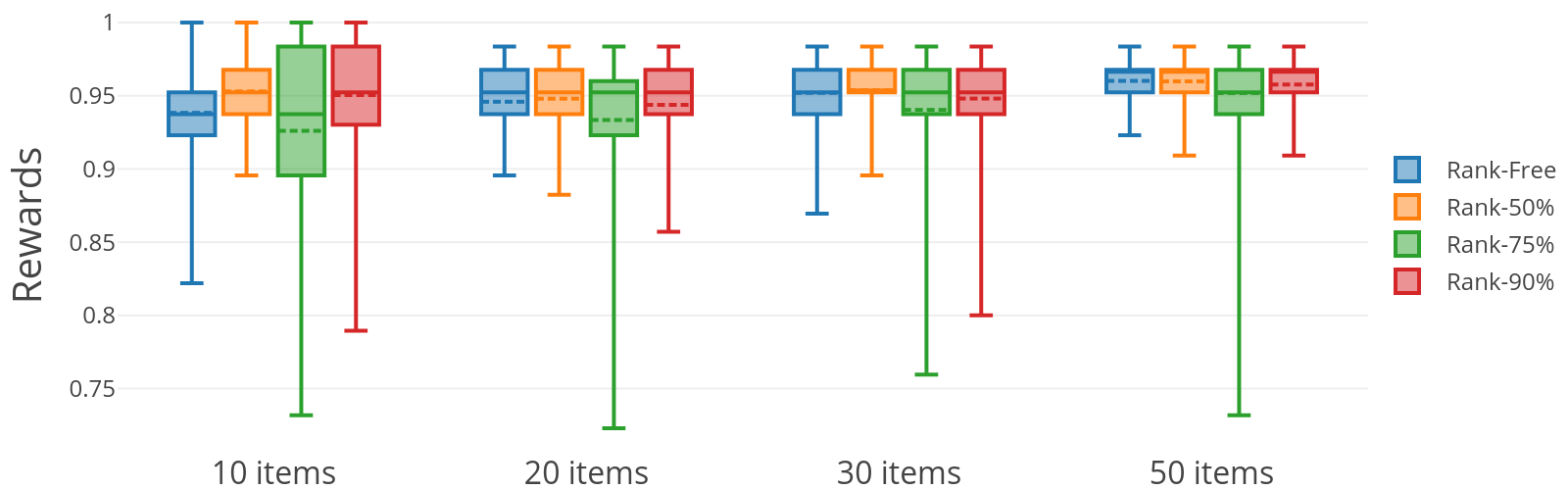}\label{fig:performance-thresholds-2d}}
\subfigure[3D BPPs with a total volume of items equals to $27000$]{\includegraphics[width=0.89\textwidth]{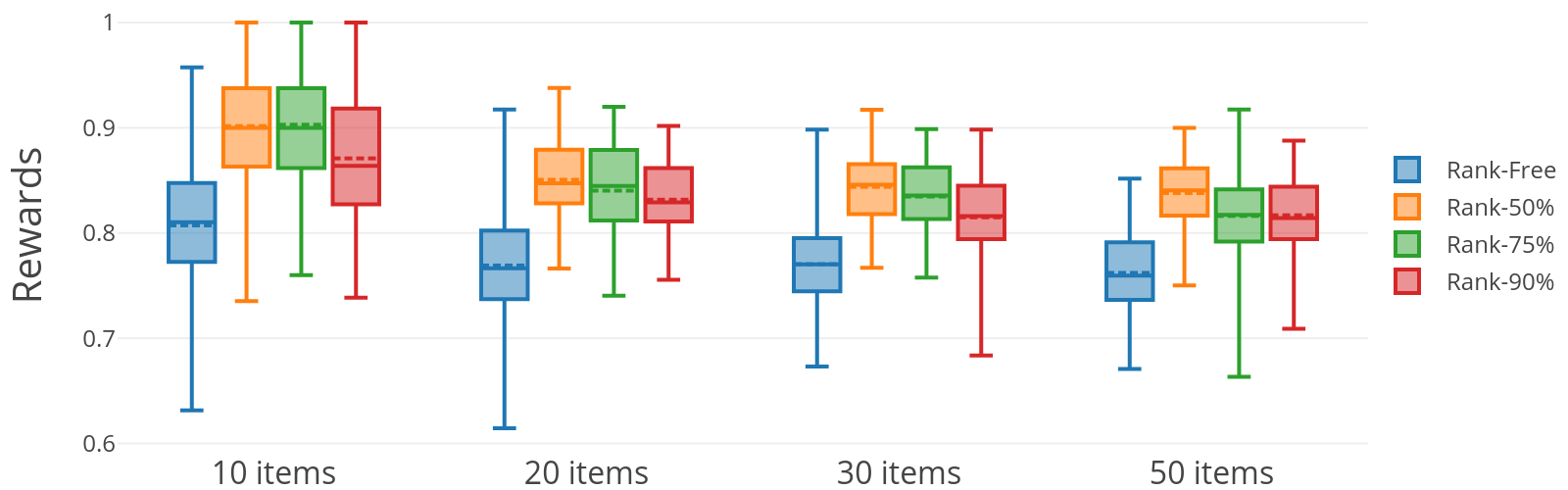}\label{fig:performance-thresholds-3d}}
\caption{Performance on 2D and 3D games for R2 networks with different percentiles.}
\label{fig:performance-thresholds}
\end{figure}

\end{document}